# Steering into New Embedding Spaces: Analyzing Cross-Lingual Alignment Induced by Model Interventions in Multilingual Language Models

**Anirudh Sundar**[1† ‡] **Sinead Williamson**[2] **Katherine Metcalf**[2]

**Barry-John Theobald**[2] **Skyler Seto**[2] **Masha Fedzechkina**[2 ‡]

[1]AI Virtual Assistant Lab, Georgia Institute of Technology

[2]Apple

asundar34@gatech.edu, {sa_williamson, kmetcalf, bjtheobald, sseto, mfedzechkina}@apple.com

## Abstract

Aligned representations across languages is a desired property in multilingual large language models (mLLMs), as alignment can improve performance in cross-lingual tasks. Typically alignment requires fine-tuning a model, which is computationally expensive, and sizable language data, which often may not be available. A data-efficient alternative to fine-tuning is model interventions — a method for manipulating model activations to steer generation into the desired direction. We analyze the effect of a popular intervention (*finding experts*) on the alignment of cross-lingual representations in mLLMs. We identify the neurons to manipulate for a given language and introspect the embedding space of mLLMs pre- and post-manipulation. We show that modifying the mLLM's activations changes its embedding space such that cross-lingual alignment is enhanced. Further, we show that the changes to the embedding space translate into improved downstream performance on retrieval tasks, with up to 2x improvements in top-1 accuracy on cross-lingual retrieval.

## 1 Introduction

Large language models (LLMs) exhibit impressive performance on a variety of tasks from text summarization to zero-shot common-sense reasoning (Raffel et al., 2020; Liu and Lapata, 2019; Bosselut et al., 2019; Richardson and Heck, 2023) and are increasingly deployed in a variety of fields ranging from health to entertainment (Singhal et al., 2025; Wu et al., 2023; Zhong et al., 2024). Despite these capabilities, to ensure that deployed LLMs align with human values, are non-toxic, and do not hallucinate, they often must be adapted post-training (Wei et al., 2024; Rodriguez et al., 2024; Ouyang et al., 2022).

Model interventions have emerged as data- and compute-efficient tools for model adaptation, whereby targeted updates are applied to model activations after pre-training (Rodriguez et al., 2024; Li et al., 2023; Rimsky et al., 2024). One such method is *finding experts* (Suau et al., 2022, 2024) which manipulates the activations of *expert* neurons responsible for encoding a broadly defined concept (e.g., a word or style of text) to steer model generations into a desired direction. This approach has been successfully used in a variety of domains, ranging from achieving gender parity (Suau et al., 2022), to reducing toxicity (Suau et al., 2024), studying geopolitical biases (Faisal and Anastasopoulos, 2023) and multilingual capabilities (Kojima et al., 2024) in mLLMs.

While model interventions successfully control model generations, prior work does not fully detail their effects on model performance. Two observations are relevant. First, model intervention methods increase perplexity on a fixed dataset post-intervention (Suau et al., 2024) meaning that the intervention introduces changes in how the model represents language. Second, prior work (Kojima et al., 2024) has shown that intervening on experts for a given language not only increases the probability of the mLLM generating text in that language but also leads to an improvement in prompt-based translation performance, suggesting that the intervention may increase the alignment between representations of different languages.

In this work, we focus on representational changes in mLLMs, with an emphasis on cross-lingual alignment, for two reasons. First, gains in mLLM performance are largely attributed to better alignment of multilingual representations (Wu et al., 2024; Lample et al., 2018). This has generated a lot of interest in improving multilingual alignment (Chaudhary et al., 2020; Efimov et al., 2023; Lample and Conneau, 2019; Liu et al., 2025). Second, datasets with the same text in multiple lan-guages are available for a variety of tasks, which enables us to study the impact of the intervention in a controlled way across multiple languages.

[†]Work completed during an internship at Apple
[‡]Equal contribution

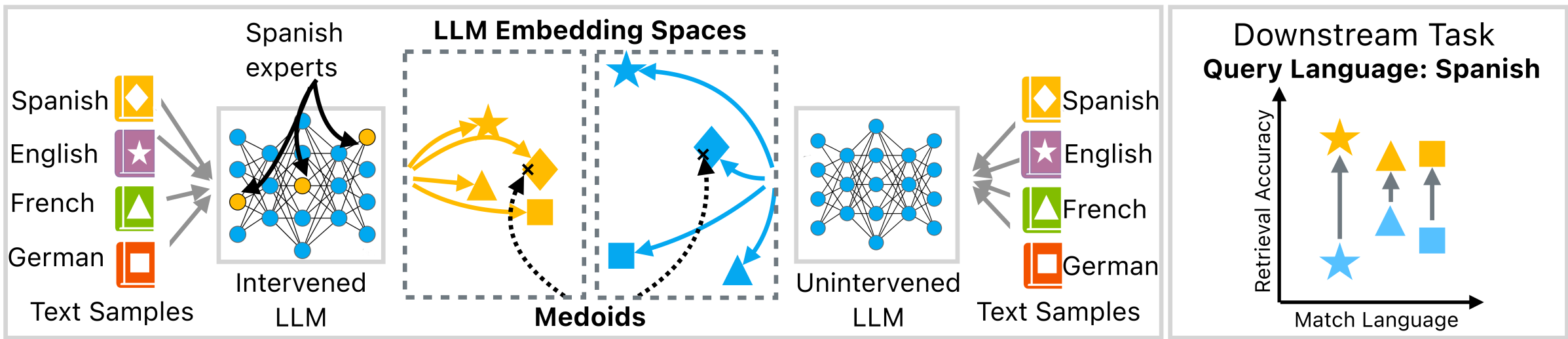

Figure 1: Following the intervention on expert neurons for Spanish, the LLM embeddings for text from different languages cluster more closely together (left, see Section 4). As a result, this intervened model is better than the unintervened model at cross-lingual retrieval where the task is to retrieve the correct translation of a sentence in a query language (right, see Section 5).

Specifically, we examine changes in the embedding space of mLLMs introduced by the finding experts intervention and link these changes to downstream task performance (see Fig. 1). We hypothesize that this intervention increases cross-lingual alignment in mLLMs and present results supporting this hypothesis. Specifically, we find that the intervention projects all languages into a new representation space within the mLLM that is characterized by new properties, some of which are desirable and some are not. On the downside, perplexity increases post-intervention, indicating a degradation in some aspects of the model's language modeling capability. However, the intervention also leads to more aligned cross-lingual representations, as evidenced by reduced distances between language embeddings (Section 4). The increased alignment translates into a performance gain on cross-lingual retrieval with up to 2x improvement in top-1 accuracy (Section 5), while preserving within-language similarity (Section 6).

## 2 Related Work

**Model interventions.** Model interventions are a family of approaches that manipulate model activations to control generations (Li et al., 2023; Turner et al., 2024; Rodriguez et al., 2024). Suau et al. (2022) propose a method to identify neurons in pre-trained transformer models that are most predictive of a particular concept (*expert neurons*) and show that setting the activations of these experts to their mean value can induce the presence of the target concept in model generations. Suau et al. (2024) find the expert neurons for toxic language and steer the LLM to generate less toxic text by dampening these neurons, while Turner et al. (2024) achieve detoxification by using a contrastive prompt. Rimsky et al. (2024) propose a method to control generations by leveraging the differences in residual stream activations between pairs of positive and negative examples. In mLLMs, Kojima et al. (2024) use this approach to produce more target language tokens in open-ended generation. However, prior work does not analyze the changes these interventions introduce in the representational space of mLLMs nor does it explore the impact of the interventions on cross-lingual alignment.

**Aligning multilingual representations in mLLMs.** Research on LLM representation alignment falls into two broad categories: 1) Improving model performance on downstream tasks via post-training methods such as prompt-based techniques (Huang et al., 2023; Tanwar et al., 2023), fine-tuning, or continuous pre-training (Zhang et al., 2023; Li et al., 2024). 2) Understanding where and how representation alignment is achieved in mLLMs. For example, Wendler et al. (2024) show that English-dominated mLLMs like Llama-2 use English as a pivot language and Zhao et al. 2024 systematically evaluate factors contributing to successful cross-lingual transfer in such models.

## 3 Methods

We seek to understand the impact of model interventions on the representational space of mLLMs with a focus on cross-lingual alignment. We consider three open-source mLLMs: Aya-8b (instruction fine-tuned) (Aryabumi et al., 2024), PolyLM-13b (chat version) (Wei et al., 2023), and Bloom-7b (base) (Scao et al., 2022). Since our aim is to draw conclusions about cross-lingual alignment, we want to make sure that we know what languages were seen in pre-training and include mLLMs for which a detailed description of pre-training datasets is available, excluding LLMs such as Mistral (Jiang et al., 2023), Llama (Touvron et al., 2023), and Gemma (Team et al., 2024). We begin by identifying and intervening on the language experts in the mLLMs and then study cross-lingual alignment in the embedding space and downstream task performance pre- and post-intervention.

### 3.1 Probing dataset construction

Following Kojima et al. (2024), we use the Flores200 dataset (NLLB Team, 2022) to find the expert neurons for a particular target language (i.e., the language specifically targeted by the intervention). Flores200 is a machine translation dataset containing short paragraphs sampled from Wikimedia[1] and subsequently translated into 204 languages by skilled human translators. We limit our investigations to the intervention on five target languages — English, German, French, Spanish, and Japanese. These languages are well represented in pre-training data of the models we are considering, ensuring the existence of expert neurons.

### 3.2 Identifying expert neurons

Expert neurons for a given language are identified following Suau et al. (2024), see Fig. 2. In this approach, a concept of interest $c$ (in our case, a particular language) is defined by a set of example sentences $N = N_c^+ + N_c^-$, where $N_c^+$ is a set of sentences that contain $c$ and $N_c^-$ is a set of sentences that do not contain $c$. The activations $\{\{z^c_{m,i}\}_{i=1}^{N}\}_{m=1}^{M}$ of every neuron $m$ in the MLP layers are obtained for inputs from both sets of sentences. The activations $z^c_{m,i}$ are used to predict $b^c = \{b^c_i\}_{i=1}^{N}$, where $b^c_i = \mathbf{1}_{i \in N_c^+}$. The expertise of neuron $m$ is then defined as the area under the receiver operating curve (AUROC) of this binary classification task, indicating the extent to which the activation of $m$ correctly predicts the presence of $c$. The requirement for a neuron to be considered an expert for a given language is that its performance as a classifier for the language is above chance (AUROC>0.5). In practice, however, we select only the top experts with the $k$ highest AUROC values (mean AUROC across all intervention targets for Aya-8b is 0.97, Bloom-7b is 0.99, PolyLM-13b is 0.83). We define **intervention** on each expert $m$ as setting the activations of this expert to its average activation $\mathbb{E}_i\{z^c_{m,i}\}_{i=1}^{N_c^+}$ for the probing set $N_c^+$, see Section 3.3.

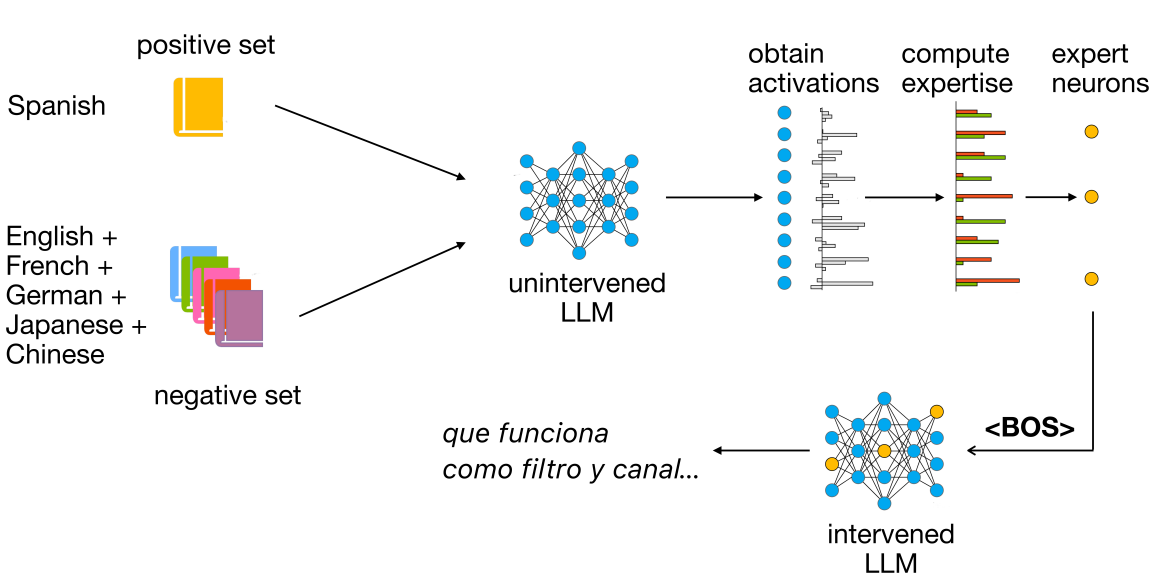

Figure 2: Illustration of the finding experts intervention. First, the activations of all MLP neurons in response to the positive and negative language examples are collected. Next, these activations are used to predict the target language label. The neurons with the highest AUROC on this task are considered experts. Intervening on the top $k$ experts increases the probability of target language generation in response to a neutral prompt.

For each of the five languages under consideration, we use the Flores200 dev split for the target language as the positive set ($N_c^+$), and the dev splits for the other four languages plus Chinese as the negative set ($N_c^-$). We include Chinese to increase variety in the character systems in the negative set but we do not consider it for the positive set ($N_c^+$).

### 3.3 Intervening on expert neurons

For the intervention, we select the $k$ neurons with the highest expertise (i.e., highest AUROC). We select the value for $k$ that balances generating text in the target language with a low perplexity on the language-specific Wikipedia text. Specifically, for each of the five languages, we sweep over expert set sizes ranging from 100 to 5000. For each setting of language and number of experts, we run free-form generation to generate 256 sentences over eight random seeds (for a total of 2048 sentences) using the beginning of sentence (<BOS>) token as the prompt. We perform generation with temperature=1 and top_p=0.9[2]. We then use lang-id (Lui and Baldwin, 2012) to measure the probability of the text generated in the target language.

To calculate the perplexity of Wikipedia text in the target language for the original and intervened models, we use the Wikimedia dump from 2023-11-01[3]. Paragraphs of text shorter than 100 char-acters are removed and the remaining paragraphs are concatenated together. Finally, a corpus of 10 million tokens is selected from the concatenated paragraphs. The context length is set to the model's maximum input size (in tokens) and a stride (i.e., the context sliding window) of 512 tokens is used to speed up the perplexity measurement. The activation of the $k$ neurons is set to their respective mean value calculated over the positive sentences (Suau et al., 2022).

[1] https://commons.wikimedia.org/wiki/Main_Page

[2] The other hyper-parameters are set to default for `transformers==4.41`

[3] https://huggingface.co/datasets/wikimedia/wikipedia

For almost all target languages, the probability of generating that language increases post-intervention (Fig. 3, top), suggesting that the intervention is successful. The only exception is English in the Aya-8b model, where the intervention reduces the likelihood of generating English. We believe that the intervention steers the model away from the default configuration, and English is the default language for that model. Interestingly, despite Bloom-7b's training set containing neither German nor Japanese, the intervention results in generating both languages with high probability. Our hypothesis is that the Bloom-7b pre-training data contains some amount of German and Japanese data that is large enough to enable expert discovery and controlled generation.

Example generations for all models are provided in App. B. Overall, all models tend to generate target language tokens. For cases where the probability of target language generation is below 1, we observe that some of the generated sentences are in a language different from the target (typically English) or contain non-language symbols such as code. We do not see generations where the tokens from different languages are mixed in the same sentence, with an exception of some Aya-8b generations where the model starts out with an English word or phrase and then continues with the target language.

While we are successfully able to increase the accuracy of target language generation through the intervention, consistent with prior work (Suau et al., 2024), we observe an increase in perplexity post-intervention as the number of activated neurons increases, see Fig. 3 (bottom), suggesting that activating experts introduces changes into the model representation. Thus, the choice of the number of experts is a trade-off between inducing the desired behavior and degrading the model. As a result, for our analyses, we set $k$ to 100 experts for Bloom-7b and 2000 for PolyLM-13b and Aya-8b.

For brevity, we present the results for the intervention on Spanish (randomly chosen) in the main text. The results for the other languages are in the respective appendices.

## 4 The intervention shifts the embedding space increasing cross-lingual alignment

We begin our investigation by quantifying the differences induced by the intervention into the embedding space. For this analysis, we intervene on each of the five target languages discussed in Section 3.3 and examine the effect of the intervention on the representations of 22 languages (the union of all languages present in the pre-training across the three language models). We exclude Arabic and Chinese from consideration due to the lack of conformity in the scripts used[4]. Note that not all of these languages are part of the pre-training for every model under consideration; however we present them for consistency (clearly indicating in all figures if the languages were seen by the model during the pre-training).

For each of the 22 languages, we embed the Flores200 test set (1012 sentences per language) with the original and intervened models' last layer. To characterize the changes in the embedding space, we calculate two types of distances: (1) the pairwise cosine distance between the embeddings of the 22 languages for the intervened and unintervened spaces and (2) the cosine distance between the mean of the embedding for each of the 22 languages and the medoid of each space (pre- and post-intervention) (Table 1).

Our findings are as follows. The intervention pulls the embeddings of all languages into a new space rather than moving them closer to the embeddings of the target language in the unintervened space (see App. C for sample visualizations of the embedding space changes using UMAP (McInnes et al., 2020) projections). The increase in perplexity post-intervention discussed in Section 3.3 also supports this finding.

The post-intervention embeddings for the different languages are closer to each other compared to the pre-intervention embeddings, as indicated by the reduced pairwise cosine distances between the languages. Specifically, the distances are reduced because the post-intervention embeddings are pulled closer to the medoid of the embedding space. As a result of the shift, all languages are closer to the target post-intervention. We notice that all distances under consideration are reduced less post-intervention for PolyLM-13b compared to the other models. We hypothesize that this relates to the specific data distribution and training procedure used for PolyLM-13b. Unfortunately, since we do not have access to the data the three models under consideration were trained on, we cannot test this hypothesis in this work. We return to this point in Section 9. Taken together, these findings suggest that the intervention projects language embeddings into a new space where they are more aligned. In the following sections, we explore if this change translates into downstream task performance.

[4]Arabic data are represented in both the Arabic and Latin scripts, while Chinese data are written using both Simplified and Traditional scripts. The decision to exclude Arabic and Chinese is motivated by prior work showing that a discrepancy in the encoding can influence performance (Blaschke et al., 2025) and several models under consideration do not provide information on which encoding was used in the pre-training. App. A contains the full list of languages and the language codes.

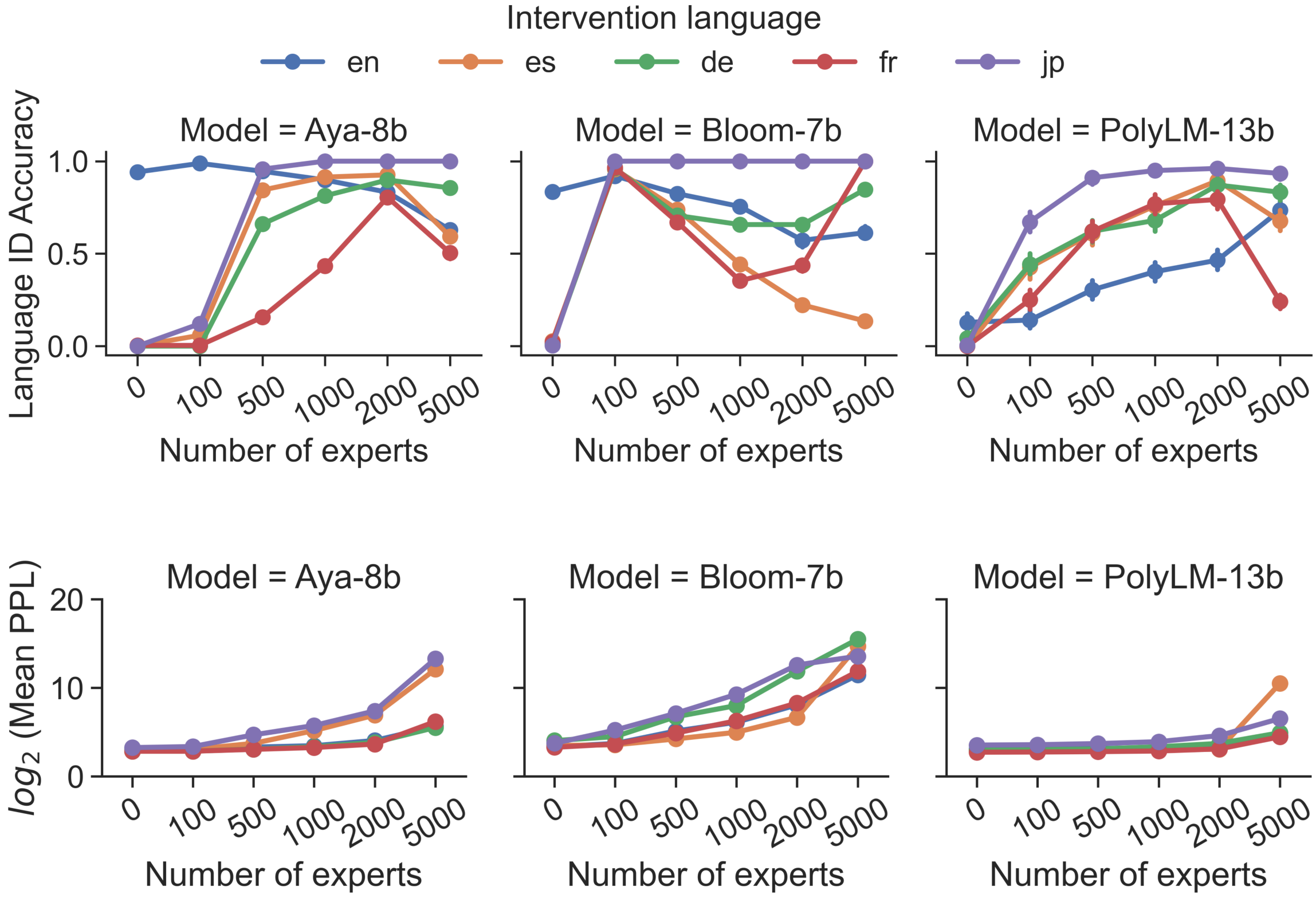

Figure 3: Language ID accuracy and Log perplexity for the intervention on five target languages. The x-axis shows the number of activated experts (0 indicates the original model). Note that German and Japanese were not in training data for Bloom-7b.

## 5 Cross-lingual retrieval performance improves post-intervention

We now ask if the increased alignment post-intervention translates to downstream task performance. We use cross-lingual retrieval as our downstream task: Given a sentence (query) in one language (query language), and a set of sentences (candidates) in a different language (candidate language), which of the candidates is a translation (match)? Our main experiments are carried out on the Flores200 test split (NLLB Team, 2022) as it allows us to test cross-lingual retrieval across multiple combinations of query and candidate languages. As the dev split of the Flores200 dataset was used to identify language experts, we also present results on the validation split of Tatoeba (Tiedemann, 2012) and the test split of BUCC-18 (Hu et al., 2020) for an independent validation of our findings (see App. H).

For each sentence, we compute pre- and post-intervention embeddings by averaging over the last hidden state of the mLLM, producing vectors with dimensions matching the model's hidden size. To identify the closest matching sentence, we compute cosine similarity between the query (e.g., in Spanish), and all candidates (e.g., in French). We select the candidate with the highest cosine similarity as the match, and then measure top-1 accuracy.

| Model | Language | Distance (all languages) | | Distance to Medoid | | Distance to Target | |
|---|---|---|---|---|---|---|---|
| | | Pre | Post | Pre | Post | Pre | Post |
| Aya-8b | Target | – | – | $0.62_{\pm 0.03}$ | $0.14_{\pm 0.03}$ | – | – |
| | Non-Target | $0.72_{\pm 0.00}$ | $0.19_{\pm 0.04}$ | $0.58_{\pm 0.01}$ | $0.12_{\pm 0.01}$ | $0.77_{\pm 0.01}$ | $0.2_{\pm 0.01}$ |
| Bloom-7b | Target | – | – | $0.60_{\pm 0.21}$ | $0.04_{\pm 0.01}$ | – | – |
| | Non-Target | $0.72_{\pm 0.00}$ | $0.17_{\pm 0.06}$ | $0.5_{\pm 0.03}$ | $0.11_{\pm 0.01}$ | $0.78_{\pm 0.03}$ | $0.13_{\pm 0.01}$ |
| PolyLM-13b | Target | – | – | $0.72_{\pm 0.04}$ | $0.43_{\pm 0.09}$ | – | – |
| | Non-Target | $0.85_{\pm 0.00}$ | $0.56_{\pm 0.09}$ | $0.72_{\pm 0.01}$ | $0.43_{\pm 0.02}$ | $0.86_{\pm 0.01}$ | $0.54_{\pm 0.02}$ |

Table 1: Cosine distances between 22 languages under consideration, mean distance to the target of the intervention, and the distance to the medoid of the embedding space are reduced post-intervention. Distance (all languages) refers to pairwise cosine distance between the embeddings of 22 languages; distance to target refers to the distance between the intervention target and the remaining 21 languages. Pre refers to pre-intervention and post to post-intervention. Distances are means and standard errors of the mean over the five intervention targets.

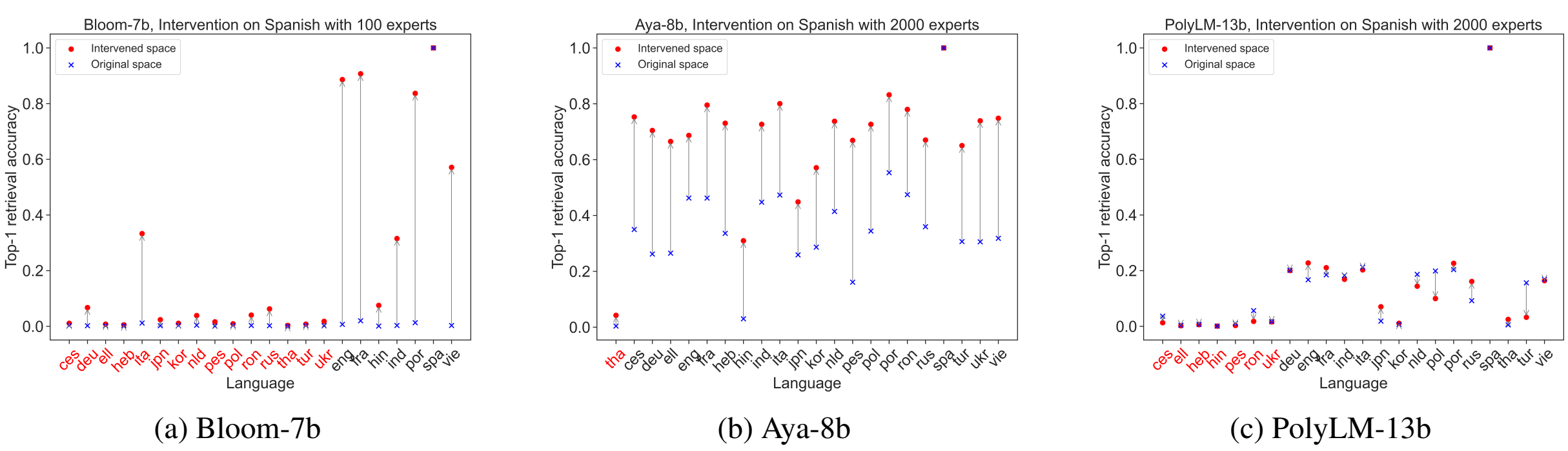

(a) Bloom-7b

(b) Aya-8b

(c) PolyLM-13b

Figure 4: Top-1 retrieval accuracy for the intervention on Spanish for 22 languages in the Bloom-7b model (left), Aya-8b (middle), and PolyLM-13b (right). The languages that are not in the training set for a given model are marked in red.

**Top-1 retrieval accuracy improves post-intervention for retrieval with the target language.** We first examine if the increased proximity to the target language in the post-intervention embedding space translates into top-1 retrieval accuracy improvement when the target is used as the retrieval query for the 22 candidate languages under consideration.

We find that top-1 retrieval accuracy improves post-intervention when using the target as the query language (see Fig. 4 for the Spanish intervention and App. E for the remaining four languages). This finding is consistent across most target languages and models. Candidate languages present in the pre-training data generally demonstrate larger gains post-intervention. The pattern of improvement differs based on the model. Specifically, for Aya-8b a successful intervention results in consistent improvements in top-1 accuracy for the majority of candidate languages (median=32%; max=74%). For Bloom-7b, top-1 accuracy gains are large (up to 89%) for a small number of candidate languages, with moderate improvements for other languages (median=14%). For PolyLM-13b, the improvements are small (median=0.5%; max=12%).

We note that accuracy gains post-intervention steadily increase with increasing $k$ activated experts from 100 to 2000 for Aya-8b and Bloom-7b, after which the performance becomes language-dependent — showing gains, drops, or no change depending on the language, suggesting diminishing returns or destabilization. We did not observe a noticeable trend in PolyLM-13b across values of $k$ over multiple languages.

To better understand how the increased alignment in the embedding space influences cross-lingual retrieval, we look at the mean pairwise cosine distances between the query and candidate languages and explore how this correlates with retrieval accuracy. Table 2 shows average correlations between post-intervention top-1 retrieval accuracy ($\text{acc}_{\text{post}}$) and mean query-candidate language distance both pre- and post-intervention ($d_{\text{pre}}$, $d_{\text{post}}$), average correlations between $d_{\text{pre}}$ and $d_{\text{post}}$, and average correlations between improvement in accuracy ($\Delta\text{acc} = \text{acc}_{\text{post}} - \text{acc}_{\text{pre}}$) and change in distance between pre- and post-intervention embeddings ($d_{\text{pre}} - d_{\text{post}}$). When calculating averages, we only include candidate languages seen in pre-training for each model; we note that the general pattern stays the same but the correlations are somewhat weaker if all 22 languages are considered for all models. We find that in this setting, when the query and intervention-target language are the same, the distance between query/target and match language is predictive of top-1 cross-lingual retrieval accuracy in both pre- and post-intervention spaces.

| Model | Query language | $r(\text{acc}_{\text{post}}, d_{\text{post}})$ | $r(\text{acc}_{\text{pre}}, d_{\text{pre}})$ | $r(d_{\text{post}}, d_{\text{pre}})$ | $r(\Delta\text{acc}, d_{\text{pre}} - d_{\text{post}})$ |
|---|---|---|---|---|---|
| | es | -0.51 [-0.88 -0.18] | -0.89 [-0.98 -0.55] | 0.48 [0.32 0.78] | 0.86 [0.49 0.96] |
| | fr | -0.64 [-0.91 -0.45] | -0.86 [-0.97 -0.57] | 0.51 [0.27 0.89] | 0.89 [0.84 0.97] |
| Aya-8b | en | -0.94 [-0.97 -0.85 ] | -0.80 [-0.96 -0.34] | 0.65 [0.44 0.92] | 0.10 [-0.69 0.60] |
| | de | -0.89 [-0.96 -0.75] | -0.87 [-0.98 -0.44] | 0.33 [-0.74 0.76] | 0.52 [0.12 0.95] |
| | jp | -0.02 [-0.62 0.34] | -0.96 [-0.99 0.34] | 0.27 [-0.18 0.62] | 0.89 [0.30 0.99] |
| | es | -0.97 [-0.99 -0.95] | -0.83 [-0.99 -0.54] | 0.79 [0.71 0.98] | 0.1 [-0.98 0.88] |
| | fr | -0.98 [-0.99 -0.94] | -0.89 [-0.99 -0.38] | 0.75 [0.62 0.99] | 0.23 [-0.98 0.83] |
| Bloom-7b | en | -0.89 [-0.99 -0.60] | -0.89 [-0.99 -0.44] | 0.97 [0.96 0.99] | 0.23 [-0.90 0.86] |
| | de | -0.90 [-0.99 -0.74] | -0.50 [-0.96 0.34] | 0.95 [0.86 0.99] | -0.72 [-0.97 0.22] |
| | jp | -0.90 [-0.99 -0.80] | NA[5] | -0.48 [-0.90 0.97] | 0.64 [-0.70 0.93] |
| | es | -0.44 [-0.91 -0.38] | -0.84 [-0.96 -0.65] | 0.70 [0.44 0.91] | 0.10 [-0.31 0.57] |
| | fr | -0.44 [-0.82 -0.35] | -0.90 [-0.99 -0.62] | 0.66 [0.20 0.93] | 0.30 [-0.18 0.82] |
| PolyLM-13b | en | -0.86 [-0.98 -0.53] | -0.84 [-0.98 -0.52] | 0.99 [0.96 0.99] | 0.28 [-0.33 0.62] |
| | de | -0.01 [-0.51 0.81] | -0.95 [-0.99 -0.57] | 0.15 [-0.56 0.57] | -0.04 [-0.71 0.34] |
| | jp | -0.52 [-0.91 0.92] | -0.96 [-0.99 0.00] | 0.73 [0.56 0.96] | 0.25 [-0.25 0.57] |

Table 2: Pearson correlations ($r$) between top-1 retrieval accuracy (acc) and mean pairwise cosine distance in the embedding space $d$. Subscripts indicate the space from which embeddings are sampled: pre = original model; post = intervened model. Numbers in brackets represent bootstrapped 95% confidence intervals. Correlations that are not statistically significant (p-values >0.05) are shown in gray.

As discussed in Section 4, all language embeddings move closer to the target's embeddings post-intervention, which explains the gains in cross-lingual retrieval accuracy. The distances in the unintervened and intervened space are positively correlated—language embeddings that are closer to the target pre-intervention are also closer to the target post-intervention. However, the magnitude of the performance gain in the intervened space does not correlate with the reduction in distance between the match and target languages across the two spaces, suggesting that the increased alignment post-intervention cannot be simply explained by a reduction in distances.

**Top-1 retrieval accuracy improves post-intervention for retrieval with the non-target languages.** In Section 4, we found that the distances between almost all languages decrease post-intervention—not just the distances to the intervention target. We next examine if these reduced cosine distances between languages *other than* the intervention target translate into improved top-1 retrieval accuracy when using these languages as the query language. For example, we study if intervening on Spanish experts improves Dutch-English retrieval (in this case, neither the query nor candidate language is the intervention-target language).

We find that, perhaps surprisingly, improvements observed when the query language is the intervention target (see Fig. 4 and Table 2) carry over to query languages other than the intervention-target language (see Fig. 5 for the Spanish intervention and App. F for the remaining four languages). For example, the intervention on Spanish expert neurons for Bloom-7b results in retrieval improvement when English, French, and Portuguese are the query language. The same intervention improves retrieval when querying Hebrew with Persian or when querying Czech with Greek in the Aya-8b model and when querying Russian with Portuguese in PolyLM-13b. These are examples of larger improvements, but many other languages follow the same pattern with smaller gains. Generally, the patterns in improvement are consistent with those seen when Spanish is the query language. Languages that are in the pre-training set have larger accuracy gains. Bloom-7b has large improvements for a small number of languages and no drops in performance. Aya-8b has relatively large improvements for a majority of languages but also has a drop in performance for some. As noted previously, PolyLM-13b performance is uneven—the improvement varies by language with languages in the pre-training set generally having larger improvements.

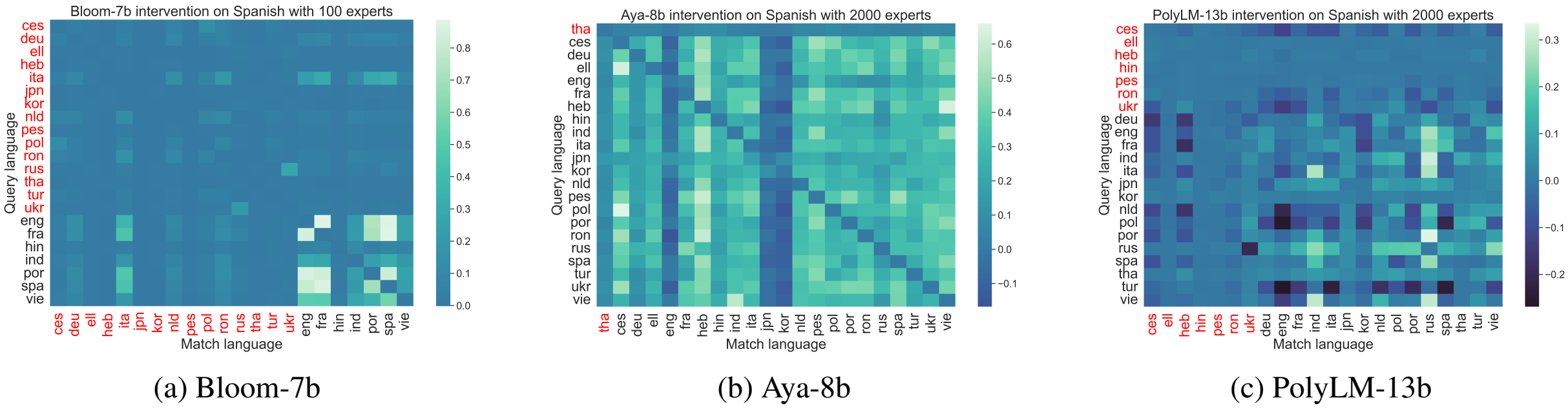

(a) Bloom-7b (b) Aya-8b (c) PolyLM-13b

Figure 5: (Top-1 $\text{accuracy}_{\text{post-intervention}}$ − Top-1 $\text{accuracy}_{\text{pre-intervention}}$) for 22 languages after intervening on Spanish expert neurons in the Bloom-7b model (left), Aya-8b (middle), and PolyLM-13b (right). The languages that are not in the training set for a given model are marked in red.

## 6 Within-language similarity is preserved post-intervention

As observed in Section 4, all languages move toward the medoid of the embedding space post-intervention, which raises the question of whether language-specific similarities are preserved in the new space. To answer this question, we evaluate model's performance on a **paraphrase** retrieval task which tests whether a sentence in the intervened space can be matched with its paraphrase in the intervened space. We utilize the PAWS-X dataset (Hu et al., 2020), which provides paired sentences across seven languages, including all five of our intervention targets. From the test split, we retain only the paraphrase pairs, excluding non-paraphrases and sentences from other languages. This transforms our evaluation into a within-language sentence retrieval task, where the goal is to match each sentence with its paraphrase from a pool of candidates for that language.

The paraphrase retrieval task reveals two key findings about embedding spaces before and after the intervention. First, the top-1 paraphrase retrieval accuracy remains largely unchanged after the intervention (see Table 3), indicating that the new embedding space preserves within-language similarity. Second, when attempting retrieval between intervened and unintervened embeddings of the same language — i.e., using the embeddings from the unintervened model as the query and the embeddings from the intervened model as candidate matches — accuracy drops significantly. This decline supports the observation that the intervention projects embeddings into a distinctly different space from their original unintervened representations discussed in Section 4. This finding also aligns with the increase in perplexity observed post-intervention – the intervened space of a given language is *not* the same space as the original space of this language.

| **Model** | **Top-1 Accuracy** | | |
|---|---|---|---|
| | (Pre) | (Post) | (Mixed) |
| Bloom-7b | 0.80 | 0.80 | 0.33 |
| Aya-8b | 0.85 | 0.86 | 0.64 |
| PolyLM-13b | 0.52 | 0.56 | 0.41 |

Table 3: Top-1 accuracy results for the paraphrase retrieval task following the intervention on Spanish. The results for other languages can be found in App. D. Pre = both the query and candidate embeddings are from the original model; Post = both the query and the candidate embeddings are from the intervened model; Mixed = query is from the original model and candidates are from the intervened model.

## 7 Intervention on random neurons does not provide an improvement on downstream tasks

In our analyses so far, we have attributed the changes in the embedding space to the interven-tion on expert neurons. Before we conclude, we consider an alternative possibility — that the expert neurons do not play a role in increasing alignment post-intervention, but instead alignment is achieved by fixing the activations of a number of neurons in a network. To address this, we assign the activation levels of the language expert neurons used in prior sections to the same number of neurons chosen randomly in the network and repeat our analyses on these models.

[5] The cosine distances between Japanese and other languages are identical in Bloom-7b in the unintervened space and thus the correlation coefficient is not defined. This is likely due to the fact that Japanese is not in Bloom 7b's pre-training.

We find that intervening on random neurons produces markedly different results compared to activating language experts (see App. G). The embedding space after the intervention on random neurons does not have the same properties as described in Section 4, which translates into the performance on downstream tasks for all models. Specifically, for the Aya-8b and PolyLM-13b top-1 cross-lingual retrieval accuracy drops for all languages post-intervention on random neurons compared to pre-intervention. Interestingly, for Bloom-7b, there is mostly no change for all target languages except French, which surprisingly improves post-intervention on random neurons. However, the gains are significantly smaller compared to those after intervening on French experts. Similar to the intervention on language experts, within-language paraphrase retrieval shows only small changes post-intervention. When they occur, these changes tend to be negative (i.e., the performance drops) after intervening on random neurons and positive after intervening on the actual language experts.

## 8 Conclusions

We present a novel analysis of the impact of the finding-experts intervention on cross-lingual alignment in mLLMs. We find that intervening on language experts projects model embeddings into a new space where languages are more aligned than in the original space but still preserve within-language similarity. These findings provide an explanation for the increase in perplexity observed post-intervention in prior work (Suau et al., 2022).

We also demonstrate that cross-lingual alignment in mLLMs can be improved through the finding-experts intervention. Applying the intervention to a single language boosts alignment across most languages seen during training and results in up to 2x improvement in top-1 retrieval accuracy. Additionally, we show that the correlation between cross-lingual alignment and cross-lingual retrieval is high and statistically significant. We recommend selecting one such language, the intervention-target, to enhance overall alignment. Our results show that this approach is most effective when the model has been trained on both the intervention-target language and the languages we aim to align.

We find that the three models we study show markedly different patterns both in the changes to the embedding space and downstream tasks. We leave it to future to work to determine the causes of these differences, though we hypothesize that they are due to the pre-training differences.

## 9 Limitations

There are several limitations that need to be considered when interpreting our results.

**We do not have access to training data or procedure.** The major limitation is that we are working with pre-trained models and we have only limited information on training data and procedure. Specifically, for Bloom-7b and PolyLM-13b, we have the information on the proportion of each language in the pre-training set. For Aya-8b, only information on which languages were seen in the pre-training (but no proportions) is available.

**PolyLM-13b is an outlier in our analyses.** PolyLM-13b emerges as an outlier in all of our analyses. We have ruled out the nature of the discovered experts as the primary factor. For example, we find that while PolyLM-13b's experts are on average lower quality (lower AUROC) than the experts in the other models, this does not fully account for its performance since we find multiple instances where PolyLM-13b and another model are matched in expert quality but PolyLM-13b still underperforms the other model. We have also rejected the hypothesis that the top experts in PolyLM-13b overlap across languages, rendering the intervention less successful — we find essentially no overlap among the top 2000 experts across five languages in any model.

Our leading hypothesis is that the discrepancy between PolyLM-13b and the other models arises from differences in training data or procedures. However, without access to training details—e.g., pretraining objectives, language distribution, data volume, or curriculum—further analysis is limited. Future work should explore the effect of intervention on alignment in a more controlled setting where the models are trained from scratch on a publicly available dataset manipulating language proportions in the training data to better understand what is driving the difference.

**We study the impact of only one intervention on alignment.** We have studied only one approach out of a family of approaches to controllable generations (Rimsky et al., 2024; Suau et al., 2024; Rodriguez et al., 2024). Each approach in the family comes with its differences – in the way the neurons targeted by the intervention are discovered, how the changes are introduced to the activations, how many neurons are intervened on, etc. We do not fully understand how these design decisions impact the representation space. For example, it is possible that some of these approaches are more beneficial for alignment while others introduce changes that are more beneficial for other tasks (or not at all). The comparison of approaches is beyond the scope of current work and we leave it for future investigations.

## A List of languages studied

The following languages are considered in this work:

| # | Language | ISO 639-1 | ISO 639-3 |
|---|---|---|---|
| 1 | Thai | th | tha |
| 2 | Czech | cs | ces |
| 3 | German | de | deu |
| 4 | Greek | el | ell |
| 5 | English | en | eng |
| 6 | French | fr | fra |
| 7 | Hebrew | he | heb |
| 8 | Hindi | hi | hin |
| 9 | Indonesian | id | ind |
| 10 | Italian | it | ita |
| 11 | Japanese | ja | jpn |
| 12 | Korean | ko | kor |
| 13 | Dutch | nl | nld |
| 14 | Persian | fa | pes |
| 15 | Polish | pl | pol |
| 16 | Portuguese | pt | por |
| 17 | Romanian | ro | ron |
| 18 | Russian | ru | rus |
| 19 | Spanish | es | spa |
| 20 | Turkish | tr | tur |
| 21 | Ukrainian | uk | ukr |
| 22 | Vietnamese | vi | vie |

Table 4: ISO 639-1 and ISO 639-3 Language Codes

## B Sample generations

Listing 1: Sample generations by the three models under consideration. 0 experts indicates an unintervened model.

```
Aya-8b with 0 experts activated:

Sample 1:
The 49-year-old, who came back to England for a second spell with
    the club in 2018, has been in charge of the first team this
    season following the departure of Garry Rowett.\n\n\"It's been an
    absolute honour and pleasure for the last two years. But I'm back
    home and I'm loving it,\" he told BBC WM.\n\n\"The biggest thing
    for me and the club is that we've done a great job. It was a
    great job at Watford and a fantastic job at Stoke.

Sample 2:
[ar]\n\u0627\u0644\u0631\u0626\u064a\u0633:
    \u0627\u0644\u0633\u064a\u062f \u0643\u0648\u0631\u062a\u064a
    (\u0643\u0631\u0648\u0627\u062a\u064a\u0627)\n[zh]
    \n\u4e3b\u5e2d\uff1a
    \u79d1\u9c81\u7279\u5148\u751f(\u514b\u7f57\u5730
    \u4e9a)\n[en]\nPresident: Mr. Kurtovi\u0107\n[es]\nPresidente:
    Sr. Kurtovi\u0107[bosnio][croata]\n[ru]\n\u041f\u0440
    \u0435\u0434\u0441\u0435 \u0434 \u0430\u0442\u0435\u043b\u044c:
    \u0433-\u043d \u041a\u0443\u0440\u0442\u043e\u0432\u0438\u0447
    [\u0411\u043e\u0441\u043d\u0438\u044f \u0438
    \u0413\u0435\u0440\u0446\u0435\u0433\u043e\u0432\u0438
    \u043d\u0430]\n[fr]\nPr

Sample 3:
import React from \"react\";\nimport { Link } from
    \"gatsby\";\nimport Image from \"./image\";\nimport SocialMedia
    from \"./social-media\";\nimport { useTranslation } from
    \"next-i18next\";\nimport { GetStaticProps } from
    \"next\";\n\nconst Page: React.FC = () => {\n const { t } =
    useTranslation();\n\n return (\n <div
    className=\"about-page-layout page-layout\">\n <div
    className=\"page-title\">\n <h1>{t(\"about.title\")}</h1>\n
    <h2>{t(\"about.subtitle\")}</h2>\n </div>\n <div
    className=\"content-wrapper\">\n <div
    className=\"about-content\">\n <div
    className=\"about-section\">\n <div className=\"about-text\">\n
    <h2>{t(\"about.about\")}</h2>\n
    <p>{t(\"about.aboutContent\")}</p>\n </div>\n <div
    className=\"about-section-image-container\">\n <Image\n
    imageClassName=\"about-section-image\"\n
    imageName=\"about-image\"\n width=\"400\"\n height=\"300\"\n

Aya-8b with 2000 Spanish experts activated:
```

```
Successful generation:
A la ocios y a los desocupados no se les oprimen las leyes, ni se
    les abrumar con impuestos y recortar sus beneficios, que es un
    derecho de todo ser humano y un deber del que lo tenga que
    ejercer, pues con ello se contribuye al progreso de la humanidad.
    No hay que intentar por ninguna preposici\'on hacer el bien
    haciendo da\~no a los dem\'as.

Code-switch generation:
If an el el caso del proyecto de Ley de Financiamiento, lo que hemos
    escuchado del gobierno es que lo har\'{a}n a el nivel de 2005 o
    incluso 2004, que ser\'{i}a una desatender de el problema", dijo
    el senador.
A nivel del Senado, los 75 miembros de el Parlamento, son 200 y se
    requieren 2/3 del Senado para lograr el qu\'{o}rum, lo que son 40
    de los que voten a favor de la Ley de Financiamiento.

Unsuccessful generation:
his is the two anoles (N el el en la anop l de P o l el el en la n i
    1 y en la n\nPublicado el 23 de julio de 2010 17 de julio de 2011
    por\nEn la secci\u00f3n de el libro de la gu\u00eda \u201cP o p a
    l i c. 1 ( 1 2 2 2 2 2 2 2 1 1 1 2 2 2 2

Bloom-7b with 0 experts activated

Sample 1:
+ \")\";\n}\n\nQStringList QgsProject::entryList( const QString
    &scope, const QString &path ) const\n{\n QStringList entries;\n\n
    const QMap<QString, QgsProjectSection>::const_iterator it =
    mProjectSections.constFind( scope );\n if ( it !=
    mProjectSections.constEnd() && it path == path )\n {\n const
    QStringList entriesForScope = it children.keys();\n for ( const
    QString &entry : entriesForScope )\n {\n entries << entry;\n }\n
    }\n else\n {\n const QMap<QString,
    QgsProjectLayerGroup>::const_iterator groupIt =
    mProjectLayerGroups.constFind( scope );\n if ( groupIt !=
    mProjectLayerGroups.constEnd() && groupIt path == path )\n

Sample 2:
and </s with the rest of the sentence. For example, the
    string&lt;&lt;b&gt;Foo&lt;/b&gt;&lt/&gt;&lt;/&gt;&lt;
    /&gt;&lt;/&gt;&lt;/&gt;&lt;/&gt;&lt;/&gt;&lt;/&gt;&lt;
    /&gt;&lt;/&gt;&lt;/&gt;&lt;/
    &gt;&lt;/&gt;&lt;/&gt;&lt;/&gt;&lt;/&gt;&lt;/&gt;&lt;/&gt;&lt;
    /&gt;&lt;/&gt;&lt;/&gt;&lt;/&gt;&lt;/&gt;&lt;/&gt;&lt;
    /&gt;&lt;/&gt;&lt;/&gt;&lt;/&gt;&lt;/&gt;&lt;/&gt;&lt;
    /&gt;&lt;/&gt;&lt;/&gt;&lt;/&gt;&lt;/&gt;&lt;/&gt;&lt;
    /&gt;&lt;/&gt;&lt;/&gt;&lt;/&gt;&lt;/&gt;&lt;/&gt;&lt;
    /&gt;&lt;/&gt;&lt;/&gt;&lt;/&gt;&lt;/&gt;&lt;/&gt;&lt;/
    &gt;&lt;/&gt;&lt;/&gt;&lt;/&gt;&lt;/&gt;&lt;/&gt;&lt;/
    &gt;&lt;/&gt;&lt;/&gt;&lt;/&gt;&lt;/&gt;&lt;/&gt;&lt;
```

```
/&gt;&lt;/&gt;&lt;/&gt;&lt;/&gt;&lt;/&gt;&lt;/&gt;&lt;/

Sample 3:
The same as my own personal,'' and to ''The same as the original''
    for other sources. In other words, the same as my own personal,
    my own personal, or the original have all been considered
    equivalent responses (Table 3).\nAlthough only 7.6% of the total
    responses to question 6 were rated as ''other'', this type of
    response, which we will term ''other'', is often in fact a
    response about the source of the information or the author(s) of
    a book (e.g., ''This book was written by . . .'' or ''This was
    from a book by . . .'').

Bloom-7b with 2000 Spanish experts activated

Successful generation:
que tiene que ver con el tema de la salud mental. Por \u00faltimo,
    se pregunta si la persona que nos atiende tiene suficiente
    experiencia y si ha seguido una formaci\u00f3n espec\u00edfica en
    ese tipo de problem\u00e1ticas. Al no tener una respuesta clara a
    estas preguntas, tendemos a rechazar ese tipo de centros o no
    utilizarlos, por ejemplo, en casos de crisis o ansiedad
    extrema.\nPor todo ello, en este curso se pretende ofrecer una
    aproximaci\u00f3n general al estado de la cuesti\u00f3n de la
    intervenci\u00f3n psicol\u00f3gica en crisis, tanto si esta se
    lleva a cabo dentro de una intervenci\u00f3n psicol\u00f3gica
    como si es de tipo asistencial

Unsuccessful generation:
no existe la cadena de comando.\nPero esto si no me deja ejecutar
    nada.\nEstoy trabajando con el Windows 10 y cuando me pongo a
    ejecutarlo me dice:\n\"procesa no se ejecuta\"\nEl problema es
    que cuando voy a ejecutar el ejecutable.exe me da un error como
    que no existe el archivo.\nLo que ocurre es que no puedo ejecutar
    nada. No quiero hacer nada y no se que
    hacer.\n-\n-\n-\n-\n-\n-\n-\n-\n-\n-\n-\n-\n-\n-\n-
    \n-\n-\n-\n-\n-\n-\n-\n-\n-\n-\n-\n-\n-\n-\n-\n-\n-
    \n-\n-\n-\n-\n-\n-\n-\n-\n-\n-\n-\n-\n-\n-\n-\n-\n-
    \n-\n-\n-\n-\n-\n-\n-\n-\n-\n-\n-\n-\n-\n-\n-\n-
    \n-\n-\n-\n-\n-\n-\n-

PolyLM-13b with 0 experts activated:

Sample 1:
, \u0437\u0432\u0456\u0434\u0442\u0438,\u0432\u0443
    \u043b\u0438\u043a\u0438,\u0440\u0456\u0437\u0430
    \u043d\u0456\u0432\u0430\u043d\u043d\u0438,\u0456
    \u0441\u0442\u043e\u0432\u0431\u0443\u0440\u0438
    \u0434\u0435\u0440\u0435\u0432.\u0426\u0456\u0434
    \u0435\u0440\u0435\u0432\u0430\u043f\u043e\u0432
    \u0438\u043d\u043d\u0456\u0431\u0443\u0442\u0438
    \u0440\u0456\u0432\u043d\u043e\u043c\u0456\u0440
```

```
   \u043d\u043e\u0440\u043e\u0437\u0440\u0456\u0437
   \u0430\u043d\u0438\u043c\u0438\u0442\u0430\u0437
   \u0430\u0433\u043b\u0438\u0431\u043b\u0435\u043d
   \u0438\u043c\u0438 ,\u0456\u0437\u0437\u0430\u043a
   \u0440\u0456\u043f\u043b\u0435\u043d\u043d\u044f
   \u043c\u0432\u0441\u0435\u0440\u0435\u0434\u0438 \u043d\u0456

Sample 2:
, \u0441 256-\u043c\u044f\u041c\u0431\u0430\u0439
   \u0442\u0430\u043c\u0438\u043f\u0430\u043c\u044f
   \u0442\u0438\u04381\u0413\u0431\u043e\u043f\u0435
   \u0440\u0430\u0442\u0438\u0432\u043d\u043e\u0439
   \u043f\u0430\u043c\u044f\u0442\u0438.\u0414\u043b
   \u044f\u0432\u0432\u043e\u0434\u0430/\u0432\u044b
   \u0432\u043e\u0434\u0430\u0431\u044b\u043b\u0438
   \u0441\u043f\u043e\u043b\u044c\u0437\u043e\u0432
   \u0430\u043d\u0438\u043d\u0442\u0435\u0440\u0444
   \u0435\u0439\u0441 USB 2.0.\n\n <p>The study by researchers at
   the University of Tokyo found that a single dose of probiotics
   could significantly reduce the inflammation associated with
   stress in the brain.</p>\n\n <p>This hotel has a total of 429
   guest rooms and 12 junior suites and 2 junior suites

PolyLM-13b with 2000 Spanish experts activated:

Successful generation:
, no lo conozco, pero por lo que he a corrido en la tele es muy
   villano y yo creo que es muy valiente\n pero el que ha corrido el
   que ha corrido en la tele no tiene nada que ver con lo nuestro\n
   pero yo creo que \u00e91 ha hecho un gran cosa por ejemplo el que
   ha corrido que ha corrido por la estepa es una parte de argentina
   donde esta el rio de la plata \n que va de estepa a argentina \n
   a un grupo de amigos que estaban trabajando en la ca mento y
   ellos estaban pasando por el de estepa a la estepa\n y ellos se
   percat que la que estabas pasando por ellos era un del rios a la
   de estepa\n y a ellos decidieron dar la vuelta y ir a nado

Unsuccessful generation:
2007.\n\n < < 2005 >\n < < < > >\n\n\n < < <
   2019\u5e7412\u670813\u65e5 \uff0c\u7531 \u81f3 \u3002 ,. .\n\n\n
   < < < > > < <\n\n . .\n\n\n < < < > > \n . \n\n < < < > >\n
   _(Nombre de la canci\u00f3n)\n\n Este es un ejemplo de
   c\u00f3digo que se podr\u00eda utilizar en JavaScript, para crear
   un bot\u00f3n con un sonido.\n ```javascript\n function
   playBtn(nombre) {\n var sonido;\n var cadena;\n sound = new
   Audio();\n cadena = sonido.Play();\n sound.currentTime = 0;\n
   sonido.loop = true;\n sonido.src = cadena;\n sonido.volume = 1;\n
   sonido.autoplay = true;\n sonido.onended = () => {\n
   sound.currentTime = 1;\n }\n }\
```

## C Sample visualizations of the embedding space changes

We project the multidimensional embeddings of the 22 languages under consideration into a two-dimensional space using UMAP (McInnes et al., 2020) to visualize how the embedding space changes. Sample visualizations for the changes to the embedding space pre- and post-intervention on Spanish are shown in Fig. 6.

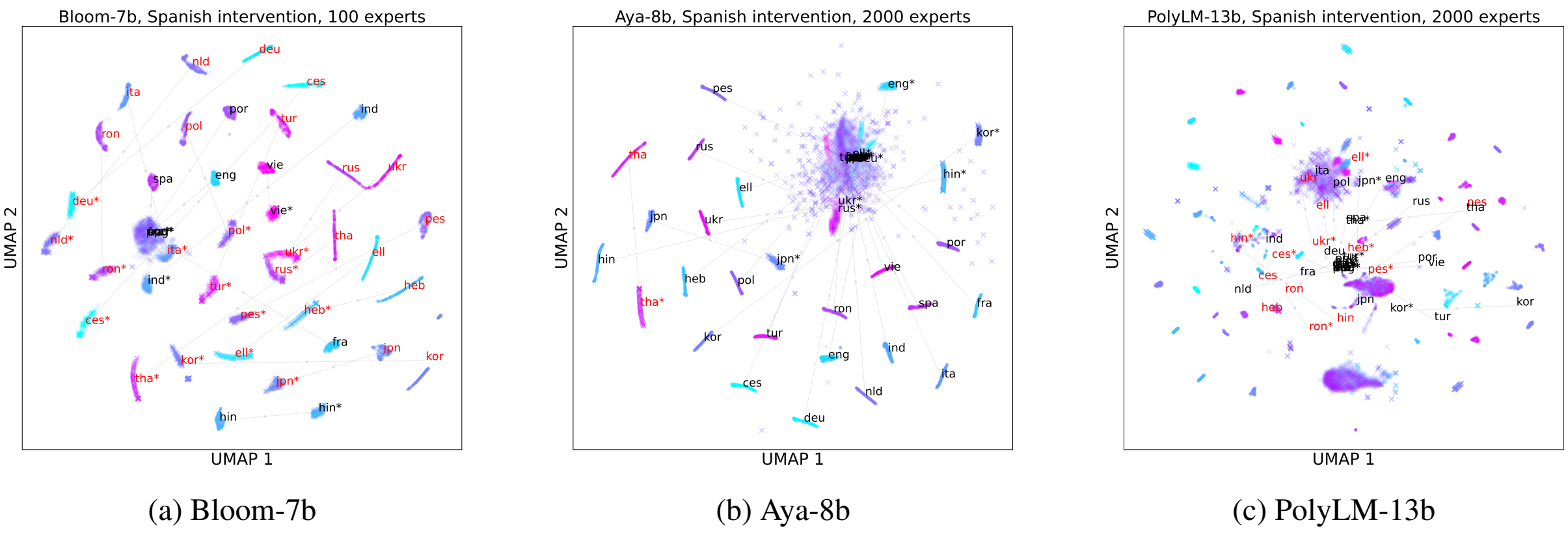

(a) Bloom-7b

(b) Aya-8b

(c) PolyLM-13b

Figure 6: UMAP embeddings for 22 languages in the Bloom-7b model (left), Aya-8b (middle), and PolyLM-13b (right). The embeddings post-intervention are marked with '*' for each language. The dots represent individual sentences in the pre-intervention space; the crosses represent individual sentences in the post-intervention space. The languages that are not in the training set for a given model are marked in red. The colors of the point clouds identify individual languages and do not carry meaning.

## D Paraphrase retrieval accuracy for four intervention-target languages

| **Model** | **Language** | **Top-1 Accuracy** | | |
|---|---|---|---|---|
| | | (Pre) | (Post) | (Mixed) |
| Bloom-7b | en | 0.80 | 0.80 | 0.71 |
| | fr | 0.80 | 0.80 | 0.26 |
| | de | 0.72 | 0.75 | 0.22 |
| | ja | 0.47 | 0.59 | 0.07 |
| Aya-8b | en | 0.87 | 0.87 | 0.56 |
| | fr | 0.83 | 0.83 | 0.75 |
| | de | 0.82 | 0.82 | 0.62 |
| | ja | 0.70 | 0.76 | 0.55 |
| PolyLM-13b | en | 0.55 | 0.53 | 0.48 |
| | fr | 0.52 | 0.50 | 0.44 |
| | de | 0.50 | 0.55 | 0.39 |
| | ja | 0.57 | 0.57 | 0.32 |

Table 5: Top-1 accuracy results for the paraphrase retrieval task for four intervention languages. Pre= both the query and the candidate embeddings are from the original unintervened model; Post= both the query and the candidate embeddings are from the intervened model; Mixed = query embedding is from the original model and the candidates are from the intervened model.

# E Top-1 cross-lingual retrieval accuracy for four intervention-target languages (query language is the same as the intervention target)

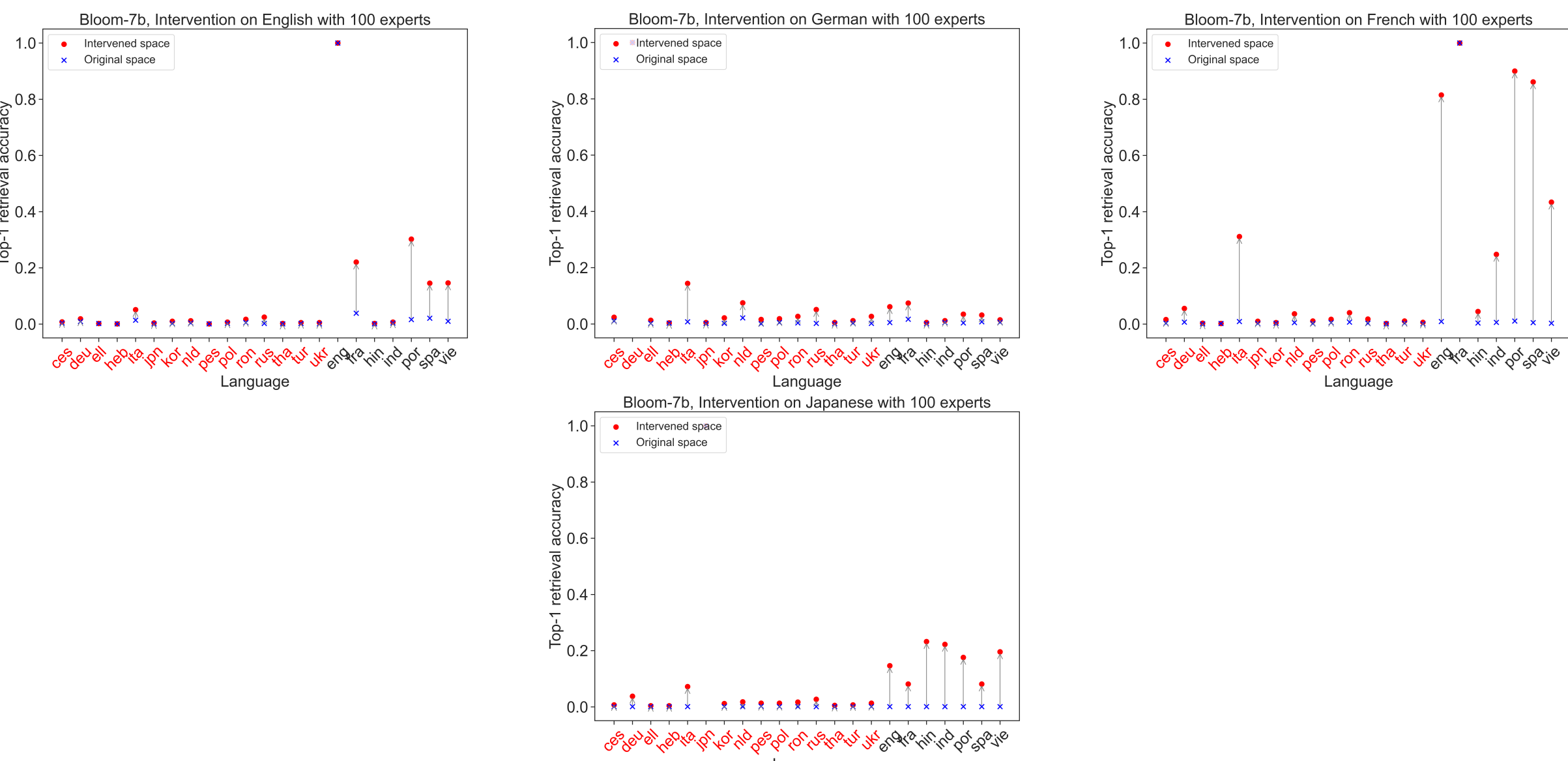

Figure 7: Top-1 retrieval accuracy for 22 languages in the Bloom-7b model. The language of the intervention is provided in the caption to each subfigure. The languages that are not in the training set for a given model are marked in red.

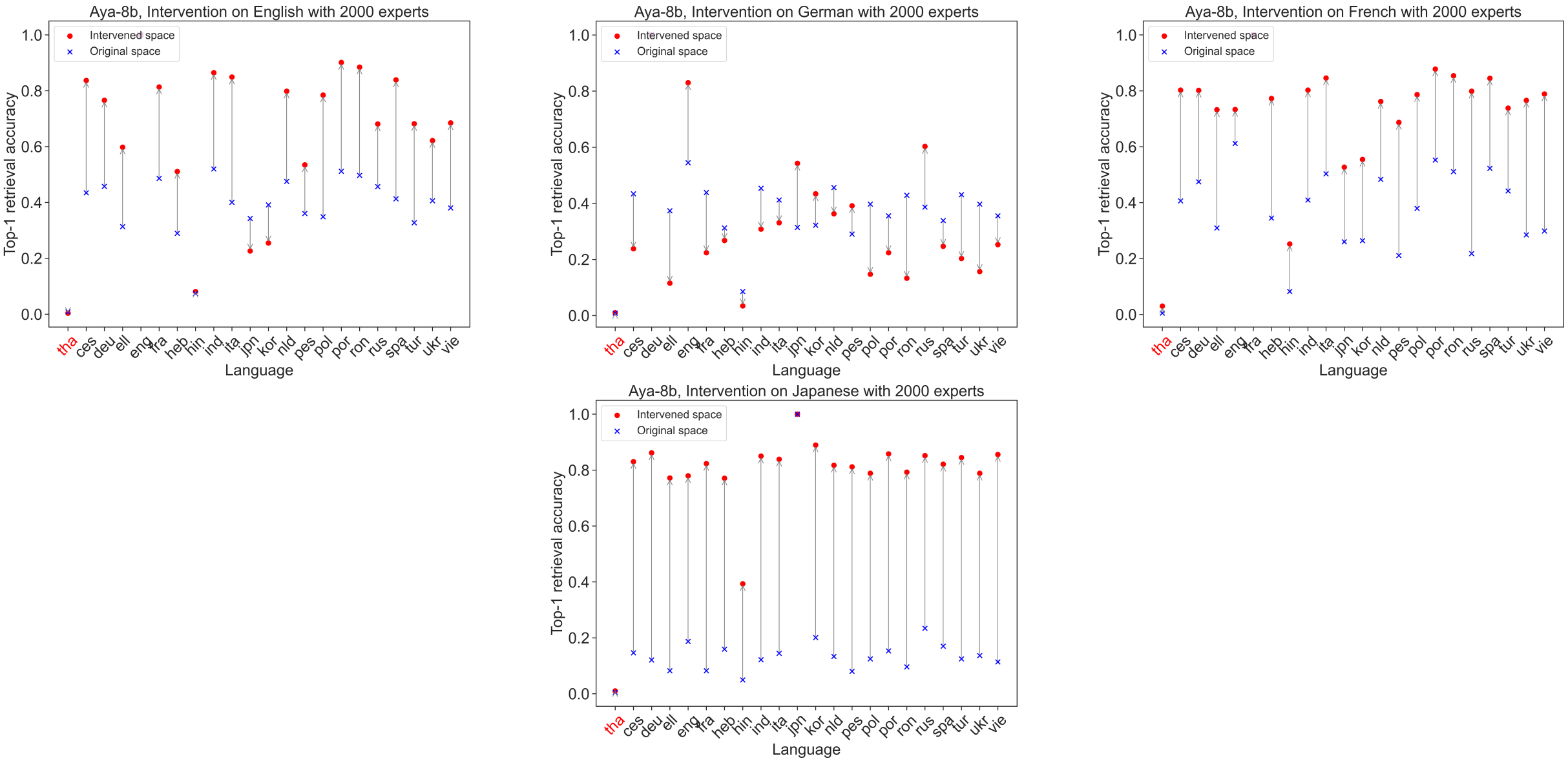

Figure 8: Top-1 retrieval accuracy for 22 languages in the Aya-8b model. The language of the intervention is provided in the caption to each subfigure. The languages that are not in the training set for a given model are marked in red.

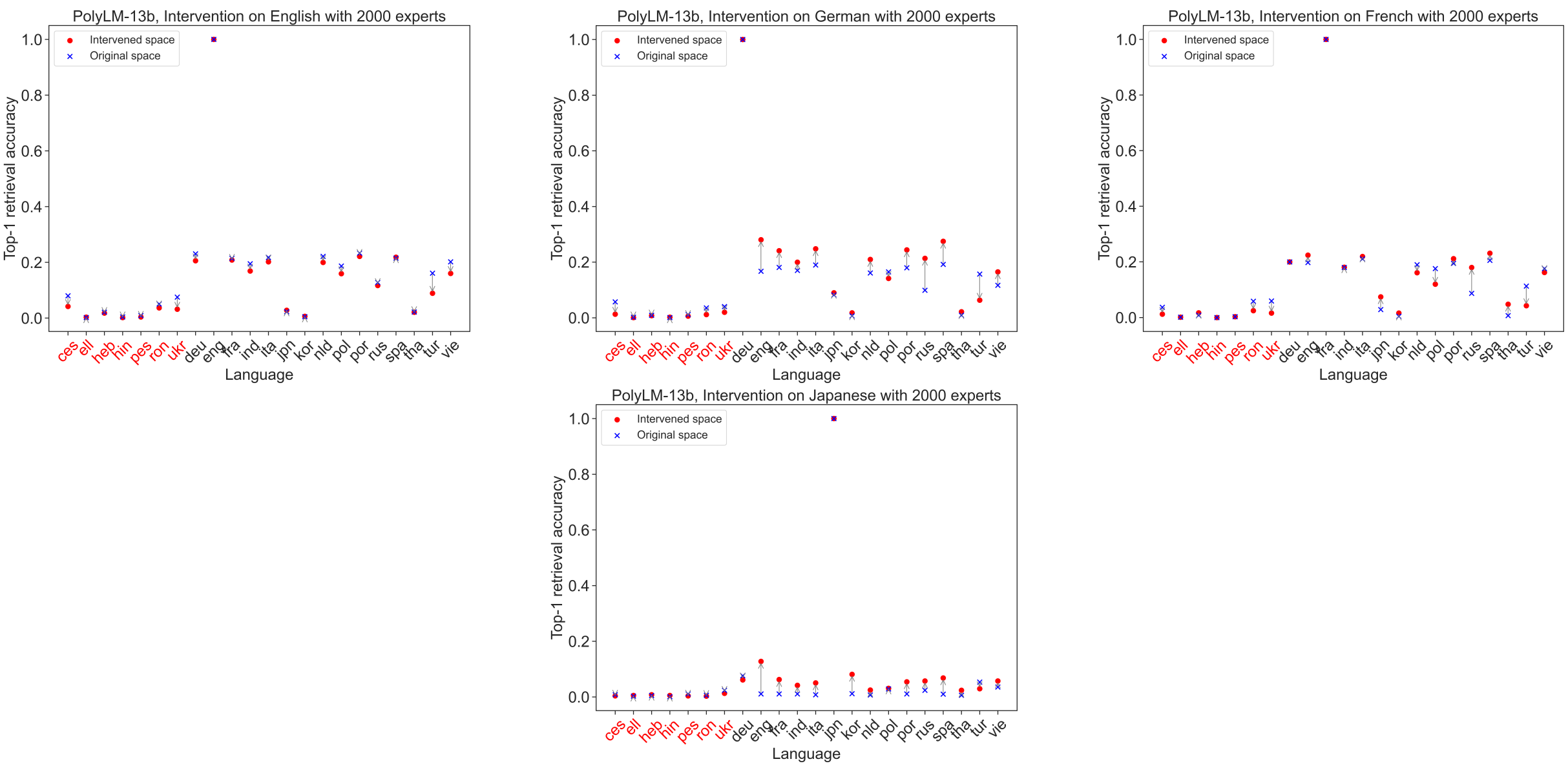

Figure 9: Top-1 retrieval accuracy for 22 languages in the PolyLM-13b model. The language of the intervention is provided in the caption to each subfigure. The languages that are not in the training set for a given model are marked in red.

# F Top-1 cross-lingual retrieval accuracy for four intervention-target languages (query language is different from the intervention target)

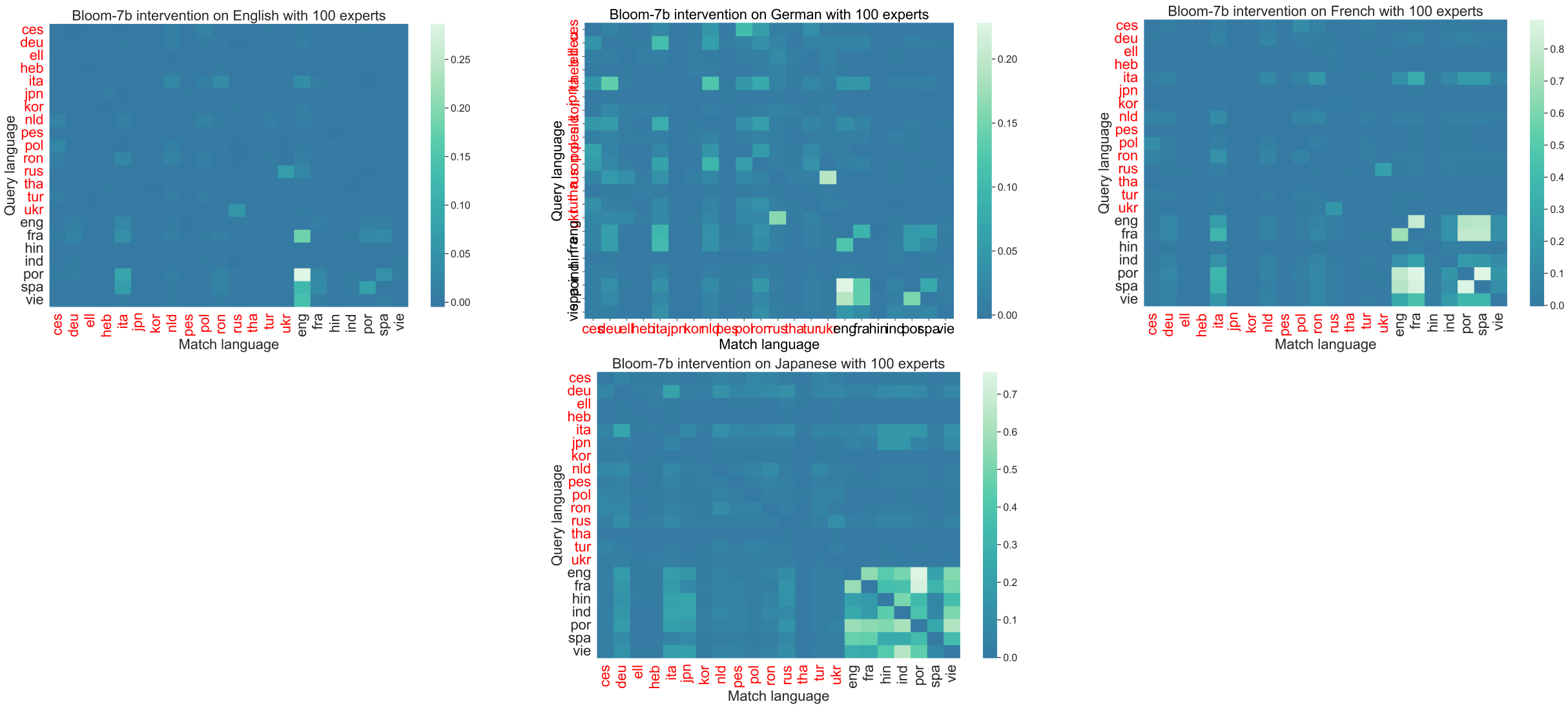

Figure 10: (Top-1 accuracy$_{\text{post-intervention}}$ − Top-1 accuracy$_{\text{pre-intervention}}$) for Bloom-7b. The language of the intervention is provided in the caption to each subfigure. The languages that are not in the training set are marked in red.

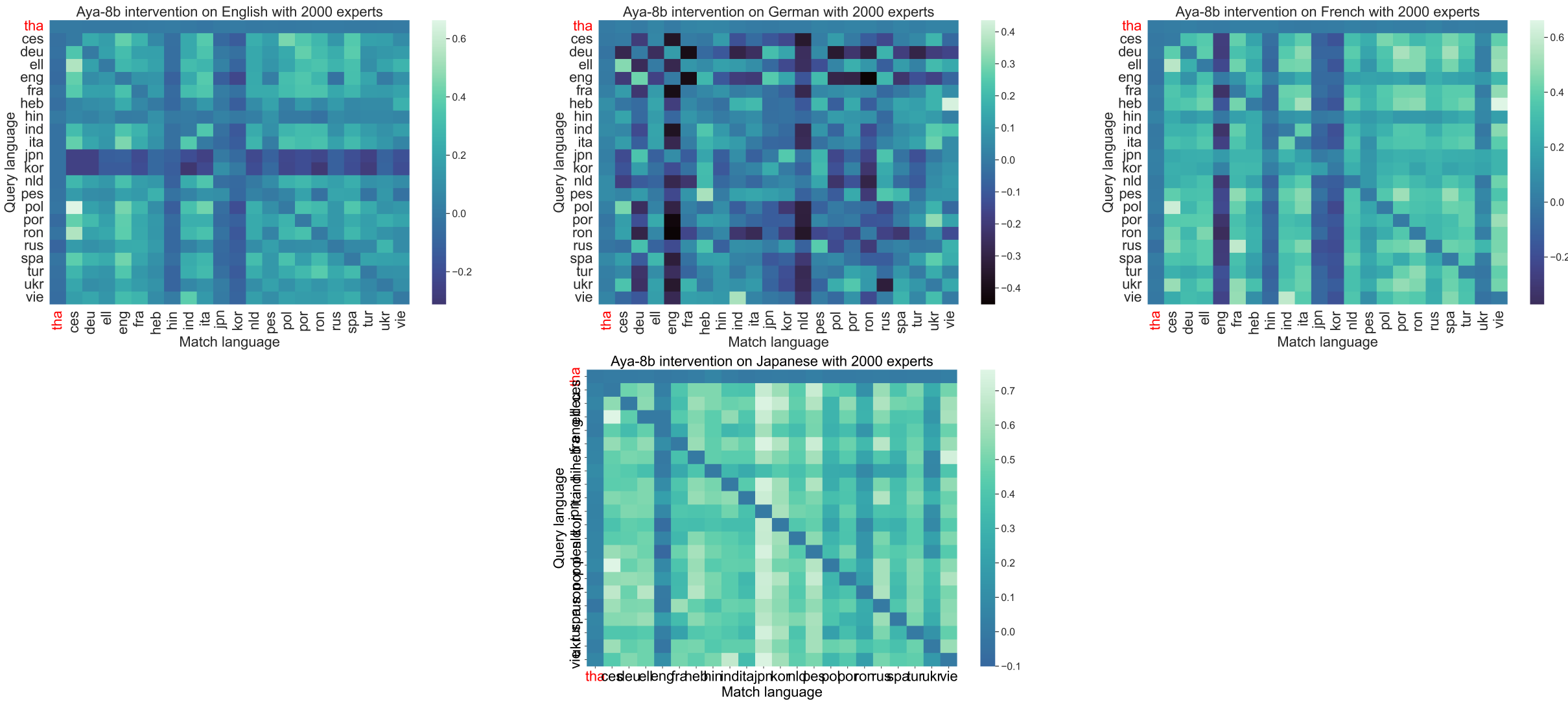

Figure 11: (Top-1 accuracy$_{\text{post-intervention}}$ − Top-1 accuracy$_{\text{pre-intervention}}$) for Aya-8b. The language of the intervention is provided in the caption to each subfigure. The languages that are not in the training set are marked in red.

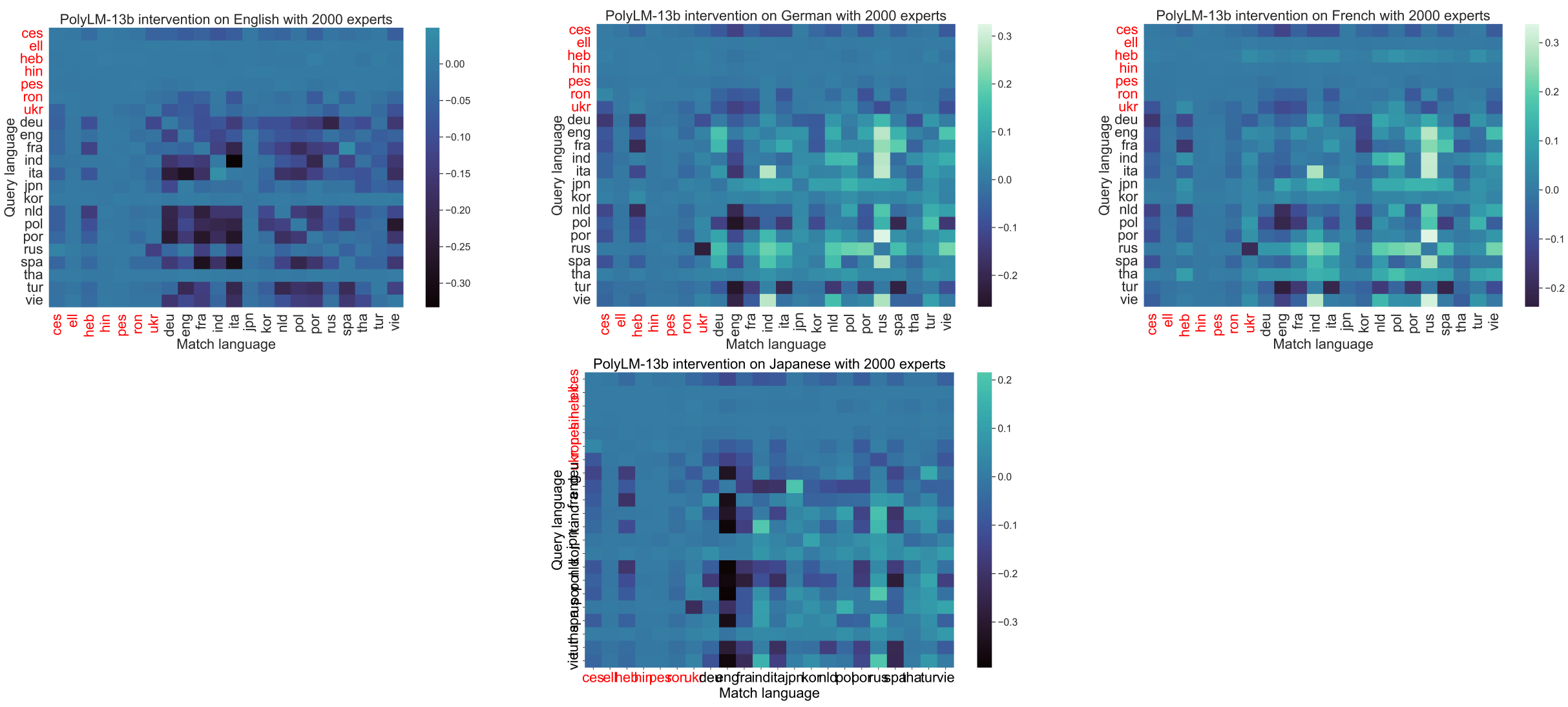

Figure 12: $(\text{Top-1 accuracy}_{\text{post-intervention}} - \text{Top-1 accuracy}_{\text{pre-intervention}})$ for PolyLM-13b. The language of the intervention is provided in the caption to each subfigure. The languages that are not in the training set are marked in red.

# G Results for the interventions on random neurons

## G.1 Top-1 paraphrase retrieval accuracy after the intervention on random neurons

| Model | Language | Accuracy | | |
|---|---|---|---|---|
| | | Pre | Post | mixed |
| Bloom-7b | en | 0.80 | 0.80 | 0.78 |
| | es | 0.80 | 0.79 | 0.62 |
| | fr | 0.80 | 0.71 | 0.00 |
| | de | 0.72 | 0.72 | 0.62 |
| | ja | 0.47 | 0.45 | 0.35 |
| PolyLM-13b | en | 0.55 | 0.55 | 0.51 |
| | es | 0.53 | 0.53 | 0.48 |
| | fr | 0.53 | 0.54 | 0.50 |
| | de | 0.50 | 0.54 | 0.45 |
| | ja | 0.60 | 0.58 | 0.23 |
| Aya-8b | en | 0.87 | 0.81 | 0.00 |
| | es | 0.85 | 0.73 | 0.01 |
| | fr | 0.83 | 0.70 | 0.01 |
| | de | 0.82 | 0.70 | 0.00 |
| | ja | 0.70 | 0.44 | 0.00 |

Table 6: Top-1 accuracy results for the paraphrase retrieval task for five intervention languages for the intervention on random neurons. Pre= both the query and the candidate embeddings are from the original unintervened model; Post= both the query and the candidate embeddings are from the intervened model; Mixed = query embedding is from the original model and the candidates are from the intervened model.

## G.2 Top-1 Retrieval accuracy for interventions on random neurons

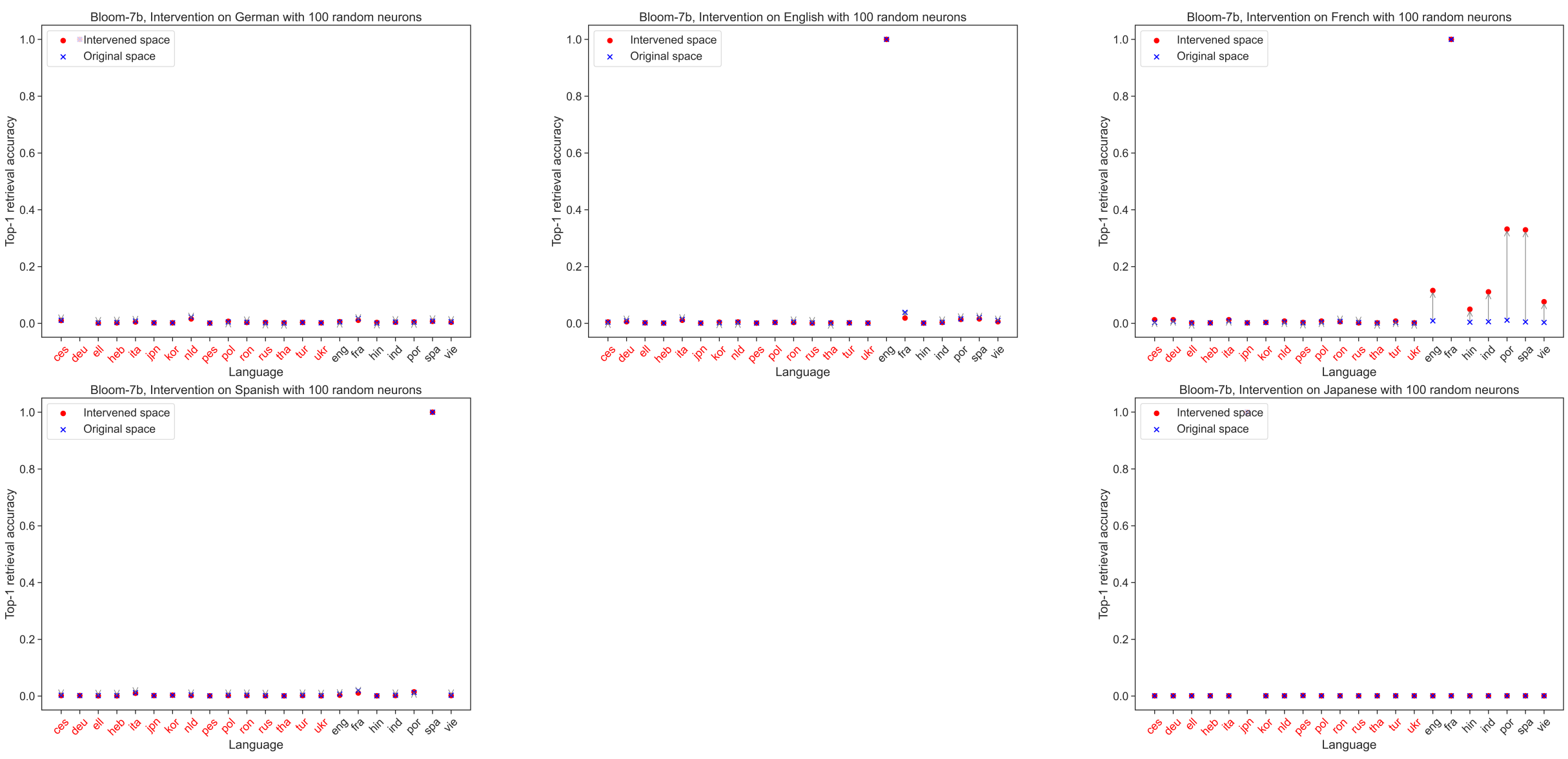

Figure 13: Top-1 retrieval accuracy for 22 languages in the Bloom-7B model with the intervention on 100 random neurons. The language of the intervention is provided in the caption to each subfigure. The languages that are not in the training set are marked in red.

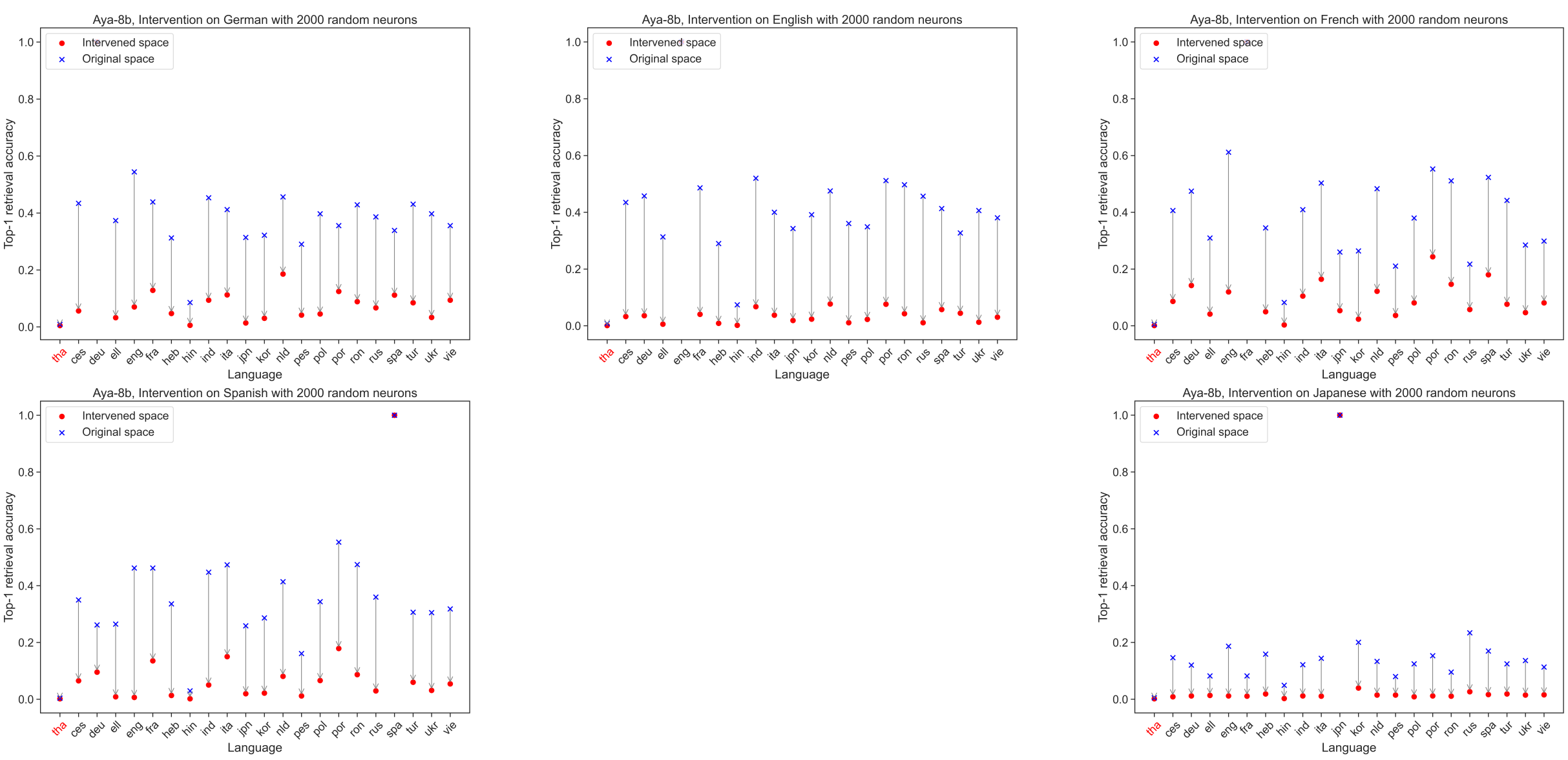

Figure 14: Top-1 retrieval accuracy for 22 languages in the Aya-8B model with the intervention on 2000 random neurons. The language of the intervention is provided in the caption to each subfigure. The languages that are not in the training set are marked in red.

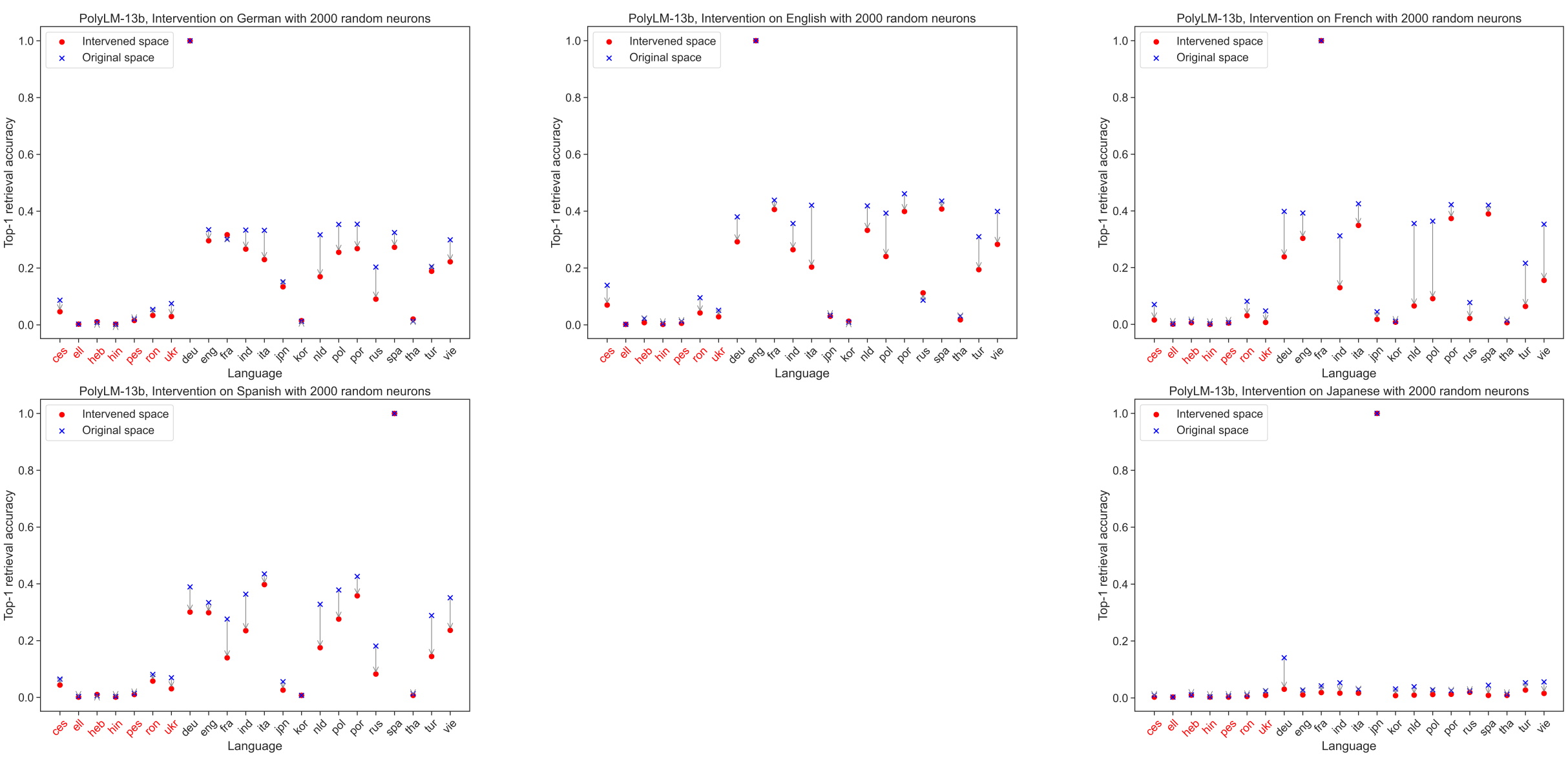

Figure 15: Top-1 retrieval accuracy for 22 languages in the PolyLM-13b-chat model with the intervention on 2000 random neurons. The language of the intervention is provided in the caption to each subfigure. The languages that are not in the training are marked in red.

## H Cross-lingual retrieval results on BUCC-18 and Tatoeba

| Model | Language | Top-1 Accuracy Pre | Post |
|---|---|---|---|
| **Tatoeba** | | | |
| Aya-8B | es | 0.114 | 0.415 |
| | fr | 0.087 | 0.251 |
| | de | 0.119 | 0.444 |
| | jp | 0.034 | 0.307 |
| Bloom-7B | es | 0.008 | 0.551 |
| | fr | 0.011 | 0.434 |
| | de | 0.006 | 0.032 |
| | jp | 0.002 | 0.043 |
| PolyLM-13B | es | 0.082 | 0.178 |
| | fr | 0.067 | 0.130 |
| | de | 0.029 | 0.171 |
| | jp | 0.000 | 0.040 |
| **BUCC-18** | | | |
| Aya-8B | fr | 0.012 | 0.073 |
| | de | 0.017 | 0.332 |
| Bloom-7B | fr | 0.000 | 0.287 |
| | de | 0.000 | 0.02 |
| PolyLM-13B | fr | 0.006 | 0.286 |
| | de | 0.008 | 0.281 |

Table 7: Top-1 retrieval for the intervetion on five target languages for Tatoeba and BUCC-18. Pre= original model; Post= intervened model.

## I Computational budget

All experiments were run on 8 A100(80GB) GPUs. The total approximate running time for 90 GPU/hours Aya-8B, 120 GPU/hours for PolyLM-13B, and 110 GPU/hours for Bloom-7B.

## J License and Attribution

All datasets used in this work are supported by public licenses. PAWS-X, Tatoeba, BUCC are part of the XTREME benchmark licensed under Apache; Flores200 is licensed under Creative Commons. The pre-trained models used in this work are also supported by public licenses Bloom-7B (RAIL 1.0), Aya-8B (Creative Commons), and PolyLM-chat-13B (Apache).